# Argument Mining using BERT and Self-Attention based Embeddings

Pranjal Srivastava
Department of Computer Science and Engineering
Delhi Technological University
New Delhi-110042, India
pranjaloct22@gmail.com

Pranav Bhatnagar
Department of Computer Science and Engineering
Delhi Technological University
New Delhi-110042, India
pranavbhatnagar2000@gmail.com

Anurag Goel
Department of Computer Science and Engineering
Delhi Technological University
New Delhi-110042, India
anurag@dtu.ac.in

*Abstract*— Argument mining automatically identifies and extracts the structure of inference and reasoning conveyed in natural language arguments. To the best of our knowledge, most of the state-of-the-art works in this field have focused on using tree-like structures and linguistic modeling. But, these approaches are not able to model more complex structures which are often found in online forums and real world argumentation structures. In this paper, a novel methodology for argument mining is proposed which employs attention-based embeddings for link prediction to model the causational hierarchies in typical argument structures prevalent in online discourse.

## Introduction

Human communication is fundamentally composed of debate and argument. With online forums increasingly serving as the primary medium for discourse and discussion, the importance of automated data processing is increasing rapidly. There are various data science techniques which proved to be successful in these natural language processing tasks. But, still there is a lot of scope of research in identifying the more complex structural relationships between concepts.

The theory of argumentation and the use of logical reasoning to justify claims and conclusions is an extensively studied field, but using data science techniques to automate the process is relatively new. A prevalent practice in the initial work in argument mining is to represent the argument structure using one or more trees or tree-like structures. This provided ease of computation as various techniques existed for tress and tree-related parsing, but arguments in the real world rarely follow the ideal system that these methods imposed.

In recent times, there have been many methods to explore argument mining in the wild using argument structures that are not required to be tree-based. Architectures like Recurrent Neural Networks, Convoluted Neural Networks, Long Short-Term Memory, and Attention-based mechanisms have allowed us to leverage contextual information in making informed machine decisions. Most recently, transformer-based architectures have given state-of-the-art performance in various Natural Language Processing (NLP) related tasks. It uses attention to boost the speed of tasks. We attempt to use the same in argument mining.

The recent trend in NLP is leveraging Transfer Learning on huge pre-trained models for better performance. Transfer Learning is a technique that was instrumental in the advancements in the domain of computer vision. It was popularized in NLP in 2018 when Google released the transformer model. Since then, transfer learning in natural language processing has aided in solving several tasks with state-of-the-art performance.

We use the Cornell eRulemaking Corpus: Consumer Debt Collection Practices (CDCP) [16], a collection of argument annotations on comments from an eRule-making discussion forum, where the argumentative structures do not necessarily form trees. We use a language representation model called Bidirectional Encoder Representations from Transformers (BERT). We also use the transformer encoder layer and generate embeddings from the encodings that capture the hierarchical relations between argument components.

## Related Work

### A. Argument Mining

Argument mining has been a problem that has attracted a lot of research interest. Moens et al. [1] attempted to identify features like n-grams, keywords, parts of speech, etc., to classify argumentative text in a collection of legal texts. Levy et al. [2] proposed a three-step approach for context-dependent claim detection. In [3], the authors presented an end-to-end approach to model the arguments and their relations in corpora. In this work, the authors proposed a pipeline of three steps: Argument detection, Argument proposition classification, and Detection of the argumentation structure. They utilized manual Context Free Grammar rules to predict a tree like relation structure between the arguments.

Stab and Gurevych [4] defined three major subroutines for effective argument mining:

1. Component Identification: Separation of argumentative spans of text from non-argumentative text for a given corpus.

2. Component Classification: Identification of the various types of argument components.

3. Argument Relation Prediction: Linking of the different parts of arguments and identification of logical dependencies.

The preliminary work in Argument Mining revolved around modeling the argument structures as trees [4, 5, 6]. These works allowed using maximum spanning tree-like parsing methods for dependency mining, enhancing the computation speed and ease. But, this failed to consider the more complex and divergent graph structures that could be found in the discourse, available on resources like online forums and discussion threads. Niculae et al. [7] proposed the first non-tree argument mining approach with a factor graph model utilizing structured Support Vector Machines (SVMs) and Bidirectional Long Short Term Memory (LSTM). Galassi et al. [8] explored the LSTMs and residual network connections to focus on link prediction between argument components. Morio et al. [9] proposed another approach that utilized Task-Specific Parameterization (TSP) to encode the sequences of propositions and a Proposition-Level Biaffine Attention (PLBA) to predict non-tree arguments with boosted edge prediction performance.

*B. BERT*

Vaswani et al. [10] first proposed the transformer architecture, with its self-attention mechanism, to counter the memory and processing cost of Recurrent Neural Networks (RNNs). Both of these architectures deal with sequential data to model global dependencies. Transformers can be trained in a highly parallelizable mode compared to the innate sequential nature of RNNs.

Devlin et al. [11] introduced Bidirectional Encoder Representations from Transformers (BERT) in a breakthrough paper. This model uses pre-train deep bidirectional representations from the unlabeled text. The pre-trained model is fine-tuned with one additional layer to create cutting-edge models for a diverse range of problems.

Reimers et al. [12] utilized BERT and ELMo (Peters et al. [13]) in an open-domain argument search to classify and cluster topic-dependent arguments, with excellent results on the UKP Sentential Argument Mining Corpus and the IBM Debater - Evidence Sentences dataset. Chakrabarty et al. [14] developed a novel approach of two fine-tuning steps on BERT, the first on a distant-labeled dataset and then on the labeled persuasive forum dataset. Their approach obtained significant improvements in comparison to other state-of-the-art techniques. Ting Chen [15] proposed an approach based on BERT and Proposition-Level Biaffine Attention (PLBA) that achieved good results.

PREREQUISITES

*A. Problem Formulation*

Our inputs will be the annotated text of a user remark, with each annotation indicating an argument. Each component of the argument corresponds to a specific span specified in the annotated text. Thus, the outputs of our model will include the argument proposition type associated with each span and the outgoing edges connecting the span to other components of the argument. Fig. 1 depicts this modeling of the problem.

*B. Embeddings*

The high-dimensional vectors can be translated into the relatively low-dimensional space known as Embeddings. Thus, it becomes easier to apply the models on the embeddings. In embeddings, the semantically similar inputs are placed together to capture the semantics of the input. Embeddings can be learned and reused across the models. The major inspiration comes from word embeddings used extensively in NLP tasks. These embeddings have been shown to capture semantic information about words and model relationships easily. We have an embedding vector and Context vector that we use to depict the elements and learn the underlying feature representations in our space.

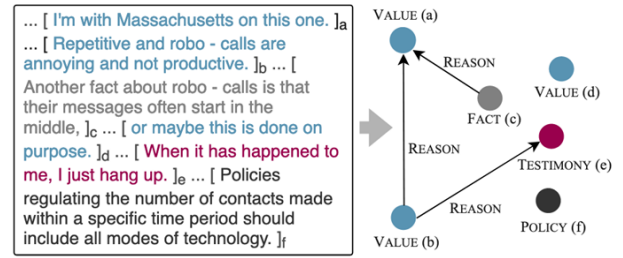

Fig. 1. Problem Formulation

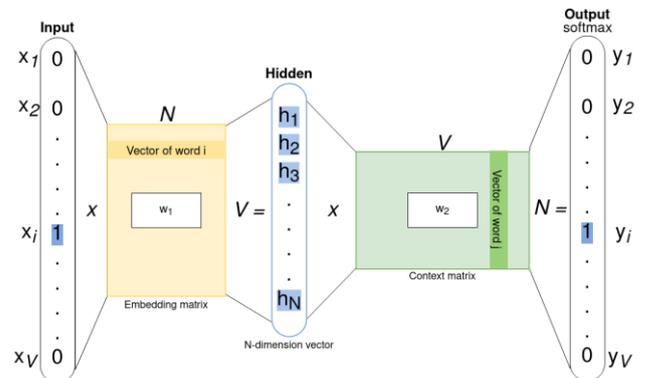

Fig. 2. Embeddings

*C. Transformer*

The transformer architecture and its self-attention mechanism were originally proposed in response to the growing computational and memory requirements of state-of-the-art Recurrent Neural Networks (RNN). By using just multi-head self-attention, transformers allow for significantly more parallel training, as opposed to the inherent sequential nature of RNN.

In our approach, we use BERT to get a sentence level embeddings for our argument components. We use these to create a sentence level representation that helps with classification as argument component type as well as downstream task of edge detection. Then, we use these encodings to create further encodings that capture context of the sentence in the argument. We use a Transformer Encoder Layer in the architecture to do so. The transformer has the following architecture as depicted in Fig. 4. The left layer is the encoding layer and the right layer is the decoding layer. We use the encoding layer in our approach.

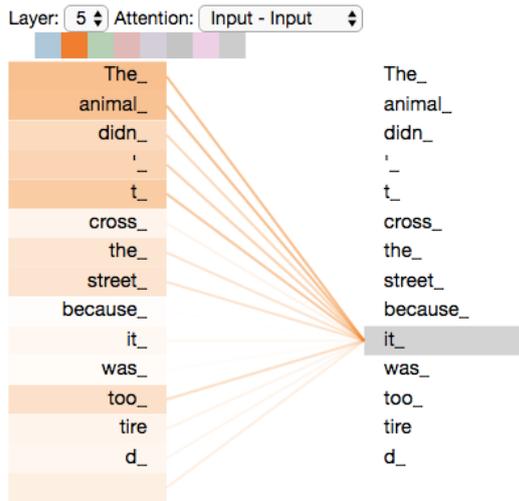

Fig. 3. Attention to context

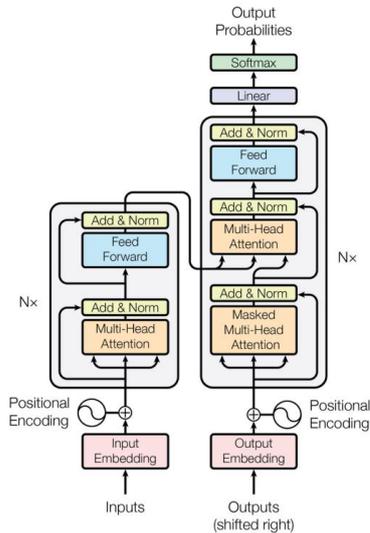

Fig. 4. Transformer

### D. BERT

In our approach to argument mining, we look rely on power of the contextual word embeddings derived from BERT. In our case we tested the 'bert-base-uncased' model from the Huggingface transformers library. BERT is essentially made up of 12 transformer encoder layers stacked on top of each other. The primary advantage of using a Transformer-based model with attention over a traditional RNN is that the attention mechanism enables the model to simultaneously see all of the words and choose which ones are most important for the given task, whereas RNN typically see words sequentially. This allows us to take advantage of computation parallelism, allowing for a model that is more efficient. BERT is pre-trained on a English corpus containing 3.3 billion words and utilising Masked Language Modelling (MLM) and Next Sentence Prediction (NSP) tasks.

**1. Masked Language Modelling (MLM)** – This model predicts a masked word in the sentence. About 15% of the words in a given input sentence are randomly masked for this task.

**2. Next Sentence Prediction (NSP)** - This model predicts whether two randomly chosen masked sentences naturally follow each other. These tasks have the unique ability to look at all the words in the sentence at once.

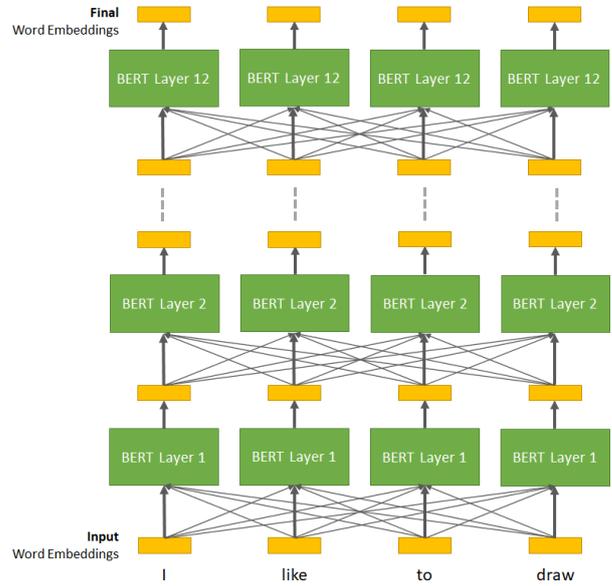

Fig. 5. BERT Architecture

PROPOSED APPROACH

The proposed approach is divided into two phases:

Phase I: Classification of argument components

Phase II: Edge detection on encodings

### I. Classification of Argument Components

In this phase, the argument spans are classified into types of argument components using BERT. The pre-trained BERT model is fine-tuned on the CDCP corpus using a supervised learning goal. Then, transfer learning is applied to the resultant pre-trained BERT model.

Since BERT produces a collection of embeddings, the classifier uses the first element of these embeddings, the special token [CLS] used to pool sentence-level semantic features. When applying BERT to a classification task, we always use the final embedding for the [CLS] token as the input to our classifier and ignore the individual token embeddings. We use the [CLS] token not just for classification but also for producing sentence-level embeddings by pooling the [CLS] output of the hidden layers for the next task of edge detection.

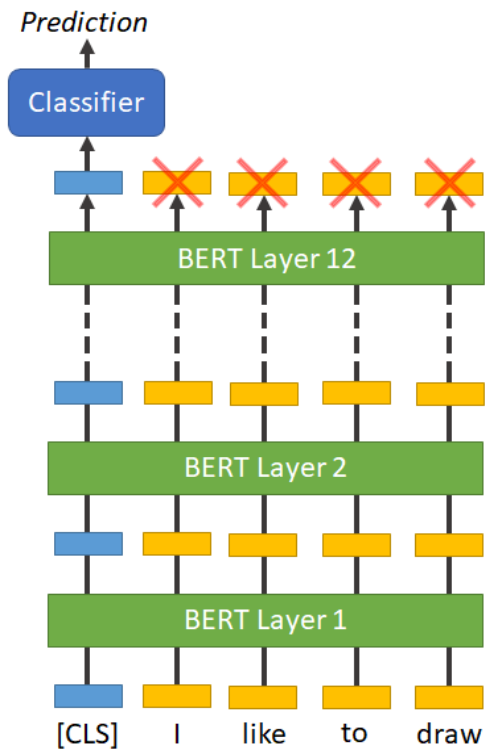

Fig. 6. Classification

## II. Edge Detection on Encodings

We take a weighted average of the last *n* layers of the hidden layer outputs to make the embeddings. This is given as:

$$e_i = \sum_{j=1}^{n} w_j * E_j$$

where $e_i$ is the sentence encoding for $i^{th}$ sentence, $w_j$ the weight for the $j^{th}$ layer and $E_j = [Emb_1 ... Emb_n]$ is the embedding of layer *j* of hidden outputs.

We pass these embeddings through a transformer encoder layer that allows a context based embedding for sentences based on the surrounding argument components. These embeddings $z_i$ are then projected to a lower dimensional space through two different projection matrices, *C* and *P* for projection into Conclusion and Projection spaces.

$$Conclusion\_emb_i = C \cdot e\_transformer_i$$
$$Premise\_emb_i = P \cdot e\_transformer_i$$

We model the following relation in the model after normalizing the projections.

$$Conclusion\_emb_i \cdot Premise\_emb_j = \begin{cases} 1, if j \Rightarrow i \\ 0, if j \not\Rightarrow i \end{cases}$$

where $j \Rightarrow i$ signifies that *sentence i* is the conclusion and *sentence j* is the premise.

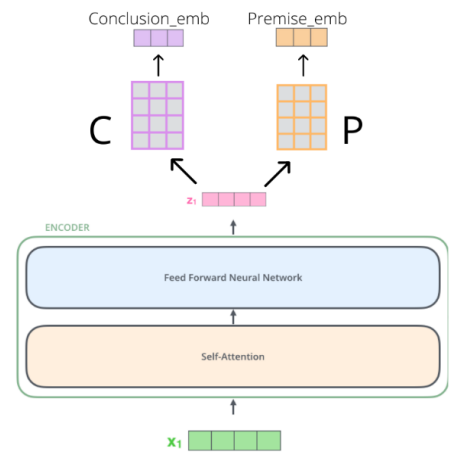

Fig. 7. Edge Detection

## EXPERIMENTATION AND RESULTS

### I. Dataset Used

For this work, we use the Cornell eRulemaking Corpus (CDCP) proposed by Park and Cardie [16], which models argument relations as links in a directed graph. It comprises about 731 user comments from an online eRulemaking forum. It has over 4500 propositions, with about 88,000 words. The propositions in the corpus are of five types:

i) Value (45%): Values are propositions that contain value judgments without specific claims or suggestions.
ii) Policy (17%): Policy is a proposition that directs a specific direction of action.
iii) Reference (1%): Reference is citing some source to support a position.
iv) Fact (16%): Fact is an assertion that can be verified with objective evidence.
v) Testimony (21%): Testimony is an objective proposition stemming from the author's experiential knowledge.

Predicting the argument structure reduces to imbalanced link prediction on the dataset, with about 3% of pairs being linked.

TABLE I. CDCP CLASSIFICATION STATISTICS

| Type | # in Training Set | # in Test Set |
| --- | --- | --- |
| **Propositions** | 3698 | 1233 |
| VALUE | 1633 | 544 |
| POLICY | 611 | 204 |
| REFERENCE | 24 | 8 |
| FACT | 592 | 197 |
| TESTIMONY | 838 | 280 |

### II. Experimental Setup

The dataset is divided into 75% and 25% as training and testing sets, respectively. Table 1 depicts the relevant split for our

     4

application. We trained the BERT model for type classification on spans of argumentative propositions. As is the norm with transfer learning, the pre-trained model has trained for task-specific optimization over three epochs using the Adam optimizer with weight decay [17] with an initial learning rate of $5 \times 10^{-3}$.

For edge detection, the last $n = 4$ layers from the BERT classification model are considered for weighted accumulation. The weights were sampled from an arithmetic progression for best results. It passes through a transformer encoder layer of dimension $768 \times 768$ which corresponds to the output dimensions of BERT, followed by projection into a 100-feature embedding space for Premise and Conclusion embeddings, respectively. We train the link prediction model for 100 epochs with AdamW optimizer with 0.4 dropout probability and early stopping for regularization.

## III. Results Analysis

We evaluate the F1 score for classification and link prediction for convenient comparative analysis. F1 score is the harmonic mean of Precision and Recall and is a suitable metric considering the imbalanced nature of precisely the link-prediction task. The results are shown in Table II. The results depict that the proposed approach outperforms all the benchmarks except the approach that utilized Task-Specific Parameterization and Proposition-Level Bi-Affine Attention (TSP+PLBA) [9]. The proposed approach gives a comparable performance with [9].

TABLE II. RESULTS

| Model | Edge Prediction | Type Prediction | Average |
|---|---|---|---|
| Deep Basic: PG [8] | 0.22 | 0.63 | 0.43 |
| Deep Residual : LG [8] | 0.29 | 0.65 | 0.47 |
| RNN : Basic [7] | 0.14 | 0.73 | 0.44 |
| SVM : Strict [7] | 0.27 | 0.73 | 0.50 |
| TSP+PLBA [9] | 0.34 | 0.79 | 0.56 |
| BERT+MLP/PLBA [15] | 0.15 | 0.86 | 0.51 |
| **Ours Approach** | 0.25 | 0.81 | 0.52 |

## Conclusion

This paper presented an attention-based approach to model causational hierarchies in typical argument structures in online discourse. The proposed approach uses BERT to get a collection of embeddings, which are then passed through a transformer encoder layer to discover edges between them. The experimental results show that the proposed approach outperforms most of the baselines and most contemporary approaches.